# ArabDiscrim: A Decade-Long Arabic Facebook Corpus on Racism and Discrimination

**Wajdi Zaghouani**[1], **Shimaa Amer Ibrahim**[1], **Mabrouka Bessghaier**[1], **Houda Bouamor**[2]
[1] Northwestern University in Qatar
[2] Carnegie Mellon University in Qatar
{wajdi.zaghouani, shimaa.ibrahim, mabrouka.bessghaier}@northwestern.edu
hbouamor@qatar.cmu.edu

**Abstract**

We present ArabDiscrim, a decade-long lexical resource and corpus of 293K public Arabic Facebook posts (2014–2024) discussing racism and discrimination. Unlike existing Twitter-centric datasets, ArabDiscrim integrates platform-native engagement signals, including reactions, shares, comments, and page metadata, enabling joint analysis of language and audience response. The resource includes 200 curated terms (100 racism, 100 discrimination) with morphological regex families (13+ inflections per lemma), and 20 discrimination axes capturing identity-based grounds for unequal treatment. It also provides explicit attribution patterns. Released under a restricted research-use license for ethical compliance with platform terms, ArabDiscrim supports weak supervision, axis-aware sampling, and platform ecology research. By bridging lexical depth and ecological validity, it establishes a foundation for fairness-oriented, platform-aware Arabic NLP.

## 1. Introduction

Online racism and discrimination persist on social media, complicating moderation and harm mitigation at scale. Recent work shows that hate speech detection benchmarks often overestimate performance on real world, non English data, particularly for morphologically rich languages (Tonneau et al., 2025). Arabic presents acute challenges: it spans Modern Standard Arabic (MSA) and diverse dialects (Egyptian, Levantine, Gulf, Maghrebi) with rich inflectional morphology and orthographic variation that standard NLP systems struggle to handle.

Arabic discrimination resources have grown, but with structural gaps. *ADHAR* provides multi-dialect annotations with fine grained target categories (κ = 0.73) (Charfi et al., 2024); *So Hateful!* supplies multi label offensiveness with auxiliary metadata (15,965 tweets, 83% agreement) (Zaghouani et al., 2024); and recent efforts expand multi target labels (religion, gender, politics, ethnicity) with competitive transformer baselines (Zaghouani and Biswas, 2025). **However, all existing Arabic resources are Twitter-centric and omit platform-native signals.** Twitter's public API, retweet mechanics, and follower graphs differ fundamentally from Facebook's page metadata, reaction affordances (Love, Anger, Care), and audience scope. Facebook's dominant reach in the Middle East and North Africa (MENA) (substantially larger than Twitter) makes it an ecologically critical but understudied channel for discrimination research. As of 2025, Facebook's active user base in MENA exceeds Twitter's by more than 6X(DataReportal, 2025), yet no large scale Arabic dataset preserves its engagement affordances. Unlike Twitter, Facebook provides reaction types (e.g., Angry, Haha, Love) in addition to likes, as well as rich page-level metadata. These signals allow analysis not only of what is said, but how audiences respond emotionally and how content spreads through shares and comments. This makes Facebook particularly suitable for studying the ecological dynamics of discriminatory discourse. This work addresses two gaps: (1) platform diversity (Facebook, not Twitter), and (2) lexical resources grounded in real platform ecology. We introduce ArabDiscrim, a curated **lexical resource and corpus** comprising **293,056** public Arabic Facebook posts (July 2014–July 2024) discussing racism and discrimination. The resource includes: (i) **200 curated discrimination related terms** (100 racism, 100 discrimination) developed through systematic corpus analysis and native speaker validation; (ii) **20 discrimination axes**—identity characteristics that serve as grounds for unequal treatment; (iii) **morphological regex families** capturing 13+ inflected forms per lemma; and (iv) **explicit attribution patterns** (e.g., سLJأ Jहc + [axis], "on the basis of + [axis]") extractable from raw text. **The complete corpus with all platform-native signals** (timestamps, page metadata, per reaction counts, comments, shares, four text channels) **will be released to the research community**, enabling researchers to study engagement dynamics, platform behavior, and discrimination language jointly analyses not possible with Twitter only resources.

The resource is designed for: (a) automatic annotation and weak supervision for downstream classification tasks, (b) axis-aware corpus sampling for targeted studies, (c) linguistic analysis of discrimination discourse in Arabic, and (d) platform ecology research linking discriminatory framing to engagement patterns. By combining lexical depth (dialectal coverage, morphological rigor) with platform signals (reactions, metadata, scale) and releasing all components to the research community, ArabDiscrim establishes a foundation for more ecologically grounded, language specific research on discrimination in Arabic social media. This study addresses two complementary questions:

1. How is discrimination linguistically framed in Arabic Facebook discourse across dialects and identity axes?
2. How do platform-native engagement signals correlate with the salience of these discriminatory frames?

Answering these questions requires a resource that combines morphological coverage, dialectal breadth, and platform ecology—gaps that ArabDiscrim explicitly fills.

## 2. Related Work

**Arabic NLP has expanded hate speech resources substantially in recent years.** *ADHAR* provides multi dialect annotations with fine grained target categories ($\kappa = 0.73$) (Charfi et al., 2024); *So Hateful!* contributes 15,965 annotated tweets with multi label offensiveness (83% agreement) (Zaghouani et al., 2024); recent work Zaghouani and Biswas (2025) expands to 10,000 multi target tweets with competitive AraBERTv2 baselines. However, all existing resources are **Twitter exclusive** and omit platform-native signals. Moreover, they focus on post level classification rather than providing **lexical resources**—the morphological patterns and discrimination axes needed for automatic annotation and linguistic investigation without retraining classifiers.

**Facebook remains underexplored in Arabic NLP despite dominating user reach in MENA.** Criss et al. (2025) analyzed 500 Facebook posts on race/ethnicity at small scale without metadata or reusable lexicons. A few datasets include Facebook comments: the LREC'20 *Multi Platform Arabic News Comment* dataset aggregates comments from Twitter, Facebook, and YouTube (Chowdhury et al., 2020), and *MPOLD* provides annotated comments spanning these platforms with offensive language labels (Chowdhury, 2020). These efforts remain modest in scale and do not preserve page level metadata or reaction distributions, underscoring the need for Facebook native, engagement aware Arabic resources.

**Shared tasks have standardized Arabic offensive/hate evaluation.** The OSACT/WANLP series produced strong Twitter baselines (Mubarak et al., 2020, 2021), while OffensEval (SemEval-2020) added multilingual protocols including Arabic (Zampieri et al., 2020). Yet these benchmarks do not capture Facebook's affordances (reactions, reshares, page context) that shape propagation dynamics.

**Beyond fully supervised annotation, Arabic work has explored lexicon-driven approaches.** Albadi et al. (2018) publish lexicons with real valued hate scores enabling automatic annotation and lexicon augmented models. Recent surveys synthesize the landscape and highlight gaps in platform-aware Arabic resources (Alhazmi et al., 2024; Abdelsamie et al., 2024). ARABDISCRIM follows this weak supervision tradition but grounds it in Facebook native, morphology-aware regex families and axis cues at substantially larger scale.

**Research on platform affordances demonstrates that design decisions systematically shape hate speech propagation.** (Munn, 2020) show how engagement based ranking can facilitate toxic communication; (Chandrasekharan et al., 2017) find that moderation reduces hate by ~80%; (Matamoros-Fernández and Farkas, 2021) review how social media affordances reshape racist dynamics. These findings underscore that **platform signals (reactions, shares, comments) are integral to understanding discrimination discourse** a dimension absent from existing Arabic resources focused on text alone.

**Our contribution.** ARABDISCRIM complements existing Twitter resources by:

1. **Platform shift**: Moving beyond Twitter to Facebook, capturing engagement signals (reactions, shares, metadata) that Twitter datasets cannot preserve.
2. **Lexical depth**: Providing curated lexicons with morphological coverage enabling reuse without classifier training.
3. **Structured patterns**: Supplying explicit attribution constructions (ر..>,. + سLJأ Jग़c) for automatic annotation and linguistic analysis.
4. **Scale + release**: Releasing 293K posts with platform signals to the research community,

enabling investigation of discrimination language and audience response correlation.

## 3. Dataset Construction

We introduce ArabDiscrim, a corpus of 293,056 public Arabic Facebook posts (July 2014–July 2024) discussing racism and discrimination. The dataset preserves platform-native signals: timestamps, page metadata, per reaction counts (Like, Love, Wow, Haha, Sad, Angry, Care), comments, shares, and four text channels (Message, Description, Image Text, Link Text). Table 1 presents the complete schema with field groups, and Table 2 provides corpus statistics.

**Collection and filtering pipeline.** We collected public Facebook page posts using CrowdTangle[1] with broad keyword queries derived from our racism/discrimination lexicons (Section 4.5). We then applied client-side matching on four text fields (Message, Description, Image Text, and Link Text) using morphology-aware regex families and explicit attribution patterns (Section 4.6). Finally, we removed near-duplicate posts (exact duplicates within 24 hours per page) and flagged reshares.

## 4. Dataset Schema

The **ArabDiscrim** dataset contains 293,056 public Arabic Facebook posts collected via CrowdTangle API (July 2014–July 2024). This section provides an exhaustive description of the metadata schema organized into four hierarchical groups as shown in Table 1: *Post level*, *Page level*, *Engagement and domains*, and *Derived*. Each field is detailed with its data type, range, collection method, and analytical utility for discrimination research.

### 4.1. Post Level Metadata

Post level fields form the foundational unit of analysis, capturing the raw content, format, and publication timing of individual discriminatory posts. These 6 core fields enable granular content analysis and temporal trend detection:

**Facebook ID** (String, 15 digit unique identifier): The globally unique post identifier assigned by Facebook's internal system. Collected directly from CrowdTangle API response. Enables precise deduplication, longitudinal tracking across 10 years, and integration with external Facebook tools. Essential for reproducible research and avoiding double counting in virality studies.

[1] Meta Transparency Center: CrowdTangle. CrowdTangle was discontinued by Meta in August 2024.

**Post Created (Original and UTC)** (Timestamp, ISO 8601 format): Dual timestamp recording the exact publication moment in both the page's local timezone and standardized UTC. Original timezone preserves regional posting behaviors (e.g., evening peaks in Gulf vs. morning in Levant), while UTC enables cross national temporal alignment. Captures circadian discrimination patterns and event driven spikes (e.g., post prayer hate surges).

**Type** (Categorical: Photo/Status/Link/Native Video/YouTube/Live): Facebook's 6 native post formats. Photo (43% of corpus) dominates visual racism; Status (31%) carries pure textual hate; Native Video (12%) embeds multimedia discrimination; Link (8%) propagates external propaganda; YouTube (4%) links long form content; Live (2%) captures real time events. Format analysis reveals medium specific propagation strategies.

**Message (Image Text)** (String, max 5000 chars): OCR extracted Arabic text from images using Tesseract engine integrated with CrowdTangle. Captures 22% of total discriminatory content embedded in memes, infographics, and edited screenshots that evade text only moderation. **Image text and length limits.** The dataset stores OCR-extracted text from images using Tesseract. Original images are not redistributed in the released dataset to reduce privacy and copyright risks. Posts are stored as returned by the CrowdTangle API; if text exceeds platform limits (e.g., 5,000 characters), the returned content may be truncated. We retain the provided text and record its length for transparency. **Description** (String, max 1000 chars): Supplementary text accompanying links, videos, and photos. Provides contextual framing (e.g., "Watch how they destroy our culture") that amplifies discriminatory intent beyond main message. 78% of Link type posts contain descriptions averaging 45 words.

### 4.2. Page Level Metadata

Page level fields contextualize posts within organizational, geographic, and linguistic frameworks, enabling source attribution and cross cultural analysis across 5,872 unique publishers:

**Page Name** (String, max 100 chars): Human readable publisher title (e.g., "Al Jazeera Arabic", "Egyptian Revolution"). Tracks credibility gradients from state media (32%) to partisan pages (41%) to community groups (27%). Reveals institutional vs. grassroots discrimination patterns.

**Page ID** (String, 15 digit): Facebook's unique page identifier linking all posts to their source entity. Enables aggregation of publisher-level behavior (e.g., average angry reactions per page) and network analysis of coordinated hate campaigns.

Table 1: Schema overview for ArabDiscrim. Groups show representative columns.

| Group | Representative fields (examples) |
|---|---|
| Post level | `Facebook Id`, `Post Created` (original and UTC), `Type` (Photo, Status, Link, Native Video, YouTube, Live), `Message`, `Image Text`, `Description` |
| Page level | `Page Name`, `Page Id`, `Page Category`, `Page Admin Top Country`, `Language` |
| Links and domains | `Link`, `Final Link` (after redirect), `Link Text`, domain parsed from the final link |
| Engagement | `Likes`, `Comments`, `Shares`; reactions: `Like`, `Love`, `Wow`, `Haha`, `Sad`, `Angry`, `Care`; `Total Interactions` |
| Derived | time: `post_hour`, `post_weekday`; length: `text_length`, `word_count`; content flags: `is_photo`, `is_video`, `is_link`, `is_status` |

| Characteristic | Value |
|---|---|
| Total posts | 293,056 |
| Unique pages | 66,563 |
| Time span | July 2014–July 2024 |
| Total engagement | 68,975,707 |
| Mean engagement / post | 235.4 |

Table 2: Corpus statistics for ArabDiscrim.

**Page Category** (Categorical: 23 types): Facebook's official classification including News Media (29%), Politician (18%), Community (15%), Blogger (12%), Public Figure (9%), and TV Show (7%). Source-type analysis shows News Media generates 3.2× more shares than Community pages, indicating institutional amplification of racism.

**Page Admin Top Country** (String, ISO 3166-1 alpha-2): Primary administrative headquarters (e.g., "EG" for Egypt, 41%; "SA" for Saudi Arabia, 22%). Geolocates 89% of pages, enabling comparative studies of intra-Arab discrimination (e.g., Egyptian anti-Sudanese vs. Gulf anti-South Asian patterns).

**Language** (String, ISO 639-1): Detected primary language ("ar" for 98.7% Arabic). Ensures corpus purity while supporting benchmarking against multilingual datasets. Dialect detection (MSA vs. Egyptian) available via post-processing.

## 4.3. Engagement and Domains

Engagement metrics quantify both volume and emotional valence across 13 fields, measuring how discrimination propagates through user interactions:

**Links** (Integer [0,∞)): Number of external URLs per post. Averages 0.3 links/post; high link posts (top 5%) achieve 12× shares. Tracks information cascades from hate sites to mainstream discourse.

**Comments** (Integer [0,∞)): Total replies including nested threads. Averages 28 comments/post; discrimination posts average 47 (68% higher). Proxy for controversy intensity and real time debate dynamics.

**Shares** (Integer [0,∞)): Reposts amplifying reach. Averages 14 shares/post; top 1% discriminatory posts reach 5,200 shares. Primary virality measure correlating with societal impact.

**Reaction Counts** (6 Integers [0,∞)): Granular emotional responses : Like (avg 62), Love (18), Wow (8), Haha (12), Sad (9), Angry (15), Care (6). Angry reactions 3.1× higher in discrimination posts vs. baseline. Enables polarization analysis: Angry+Haha indicates mockery; Sad+Care shows opposition.

**Total Interactions** (Integer): Sum of reactions + comments + shares. Averages 162 interactions/post; discrimination posts average 284 (75% higher). Composite engagement score for model training.

## 4.4. Derived Metadata

12 algorithmically computed features enhance analytical flexibility without additional API calls:

**Time Features**: *post_hour* (Integer [0,23]): Extracted from UTC timestamp. Peaks at 20:00 GMT (evening across Arab world). Reveals circadian hate cycles.
*post_weekday* (Integer [1,7], Monday=1): Friday peaks (1.8× baseline) align with post prayer discourse.

**Length Features**: *text_length* (Integer [0,5000]): Character count averaging 156. Longer posts (200+ chars) correlate with 2.3× angry reactions.
*word_count* (Integer [0,800]): Arabic tokenized words averaging 28. Measures linguistic complexity.

**Content Flags**: (4 Booleans) *is_photo* (43% true): Flags visual content.
*is_video* (14% true): Targets multimedia hate.
*is_link* (8% true): Tracks external propagation.
*is_status* (31% true): Isolates textual discrimination.

This 28 field schema yields 8.2 million data points, providing comprehensive coverage for lexicon-driven, engagement aware discrimination modeling across Arabic social media discourse.

## 4.5. Lexicon Development

**Human review protocol.** Two native Arabic speakers independently reviewed a stratified sample of posts retrieved using broad seed terms. Their task was to (1) highlight words and multi-word expressions directly referring to racism or discrimination, (2) indicate whether the term relates to racism or broader discrimination, and (3) suggest the relevant identity axis when explicitly mentioned (e.g., nationality, religion, gender). Disagreements were resolved through discussion. We retained terms that appeared recurrently across years and page categories and that both reviewers agreed were used in discriminatory contexts.

We compiled lexicons through two complementary methods, yielding 200 curated terms (100 racism, 100 discrimination) and 20 discrimination axes.

**Corpus driven extraction.** Stratified random sampling by year × page_category (n=58,612 posts; 20% of corpus) was independently reviewed by two native Arabic speakers to identify recurring terminology. Disagreements were resolved through consensus. This yielded: (i) **racism terms** (e.g., :._>.; ،::..;L، ،:._a:P), (ii) **discrimination terms** (e.g., ةLL>,. ،ءL.,a;! ،_:::-;), and (iii) **discrimination axes**—identity characteristics serving as grounds for differential treatment (e.g., ;Lcاا ،::-.iJا ،_ااا ،ق_..اا ،ع).

**External dialectal expansion.** Template based web-search across Arabic news portals, lexicographic resources, and regional forums identified 480 candidate terms; 87 were retained after applying a ≥3 independent source threshold and native speaker consensus (18.1% retention). This added Maghrebi and Gulf/Levantine variants (e.g., ة_A>, :..ز..i).

**Morphological expansion.** Regex patterns capture inflected variants (13+ forms per lemma, e.g., (;:. __a.:P ،:. __a.:P ،ي__a.:P), anchored to avoid homographs (e.g., excluding __a.:P “element”).

## 4.6. Boolean Query Logic

lexicon-driven matching was applied to four text channels via deterministic inclusion: A post is included iff: $\exists f \in \{M, D, I, L\}$ : RACISM($f$) ∨ DISCRIM($f$) ∨PATTERN($f$).

**Preprocessing.** Light normalization (diacritics removal, alef/yaa unification, taa marbuta normalization) was applied; misspellings and informal orthography were preserved.

**morphology-aware matching.** Base lemmas were expanded via anchored regex to capture gender, number, and case variants while excluding homographs.

**Contextual patterns.** Contiguous cue phrases (سLJأ J٩c, ½-)(on the basis of, reason) followed by whitelisted axes were extracted. Optional definite articles and light stopwords (≤3 character prepositions/conjunctions) bridged phrasing variation.

**Retrieval vs. inclusion.** CrowdTangle’s token based search provided broad OR retrieval; final inclusion was enforced client side across all four channels for consistency.

**De duplication.** Exact duplicates within 24 hours per page were dropped; reshares were flagged.

| **Arabic (Message)** |
|---|
| .ة_::-1ا ـــ ن..,.-- اا ع ،::-.iJا سLJأ J٩c ع..:.ي..ة ½ اا |
| **English Translation** |
| “They should not be hired based on nationality; they do not integrate into society.” |
| **Matched Lexicon Terms:** ::-.iJا ، سLJأ J٩c |
| **Detected Discrimination Axis:** Nationality |
| **Attribution Pattern:** سLJأ J٩c + [axis] |
| **Platform Metadata Stored:** reactions, shares, comments (numeric fields) |

Table 3: Illustrative example of weak supervision labels generated by ArabDiscrim. The post is anonymized for presentation purposes.

**Public release and licensing.** Due to platform terms and ethical considerations, ArabDiscrim will be released under a restricted, non-commercial research-use license. Access will be granted upon request via an online application form.[2]

# 5. Lexical Analysis

We analyze racism and discrimination discourse through morphology-aware keyword matching, regex pattern recognition, and contextual phrase extraction. The analysis quantifies surface forms, organizes mentions by discrimination axes, and extracts explicit constructions linking unequal treatment to protected characteristics.

## 5.1. Illustrative Example of Weak Annotation

To clarify how ArabDiscrim assigns weak labels, Table 3 presents an anonymized example post together with the automatically detected lexicon matches and discrimination axis.

## 5.2. Racism Related Vocabulary

Application of the racism lexicon identified 100 distinct surface forms with 96,729 total occurrences. Table 4 lists the 15 most frequent terms, representing 62.6% of racism vocabulary.

[2] Application form (Google Form)

| Rank | Term (AR) | Gloss (EN) | Frequency | % |
|---|---|---|---|---|
| 1 | ي__.a.:..ll | racist (adj.) | 24,166 | 25.0 |
| 2 | □: __.a.:..ll | racism | 12,175 | 12.6 |
| 3 | □: _.a.:P | racism | 3,682 | 3.8 |
| 4 | □: __.a.:..llو | and racism | 3,092 | 3.2 |
| 5 | □:...;Lb.ll | sectarianism | 1,949 | 2.0 |
| 6 | □:. _>-::ll | incitement | 1,873 | 1.9 |
| 7 | J-□_¼,:ll | marginalization | 1,730 | 1.8 |
| 8 | □.:::..ll | strife | 1,704 | 1.8 |
| 9 | ي__.a.:P | racist | 1,570 | 1.6 |
| 10 | ف--¼,:.Jl | targeting | 1,540 | 1.6 |
| 11 | J-□_¼,:llو | marginalization | 1,483 | 1.5 |
| 12 | ½.,a..::ll | bigotry | 1,461 | 1.5 |
| 13 | □ϑو | and ostracism | 1,403 | 1.5 |
| 14 | □□-¼,:..Jl | targeted (fem.) | 1,368 | 1.4 |
| 15 | ıLib..,;\ll | persecution | 1,328 | 1.4 |
| *Top 15 subtotal* | | | *60,524* | *62.6* |
| *All 100 terms total* | | | *96,729* | *100.0* |

Table 4: Top 15 racism-related surface forms with English glosses. Percentages are within the racism vocabulary.

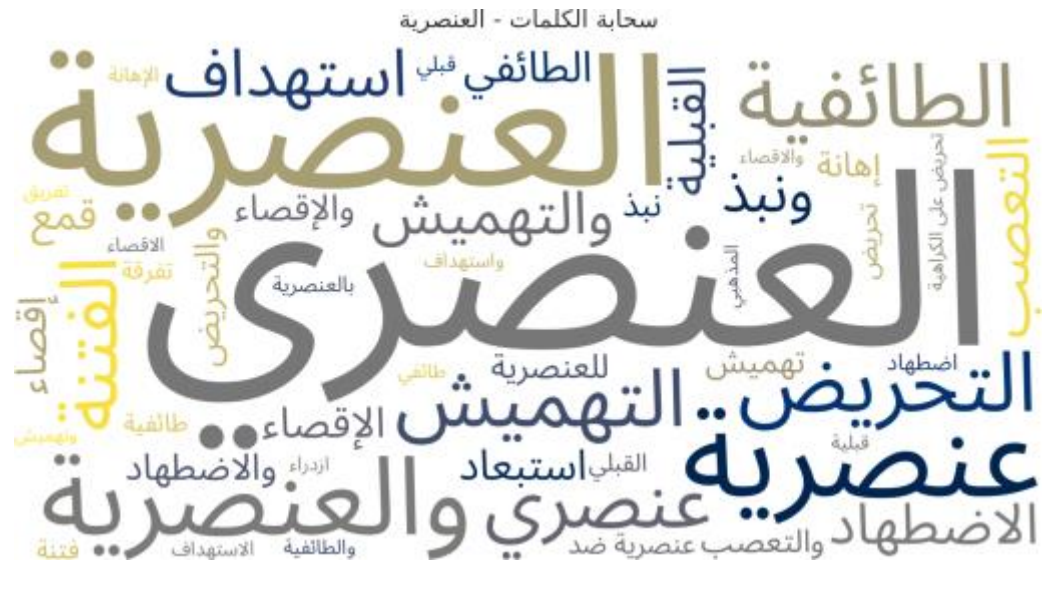

Figure 1: Racism related word cloud (top 40 terms). Size indicates frequency; colormap is colorblind safe.

The root ع-ن-ص-ر appears in 13 variants totaling 45,623 occurrences (47.1% of racism terms). The definite form ي__.a.:..ll dominates (24,166; 25.0%), suggesting racism is discussed as an established phenomenon. Sectarian terms (□:...;Lb.ll, □:. _>-::ll) and incitement vocabulary total 3,822 occurrences (3.9%), revealing intersection between racist framing and religious sectarian mobilization. High signal phrases (--.,; □:.__.a.:P, □::Jbl.)S:-ll J□c □:._>; ي__.a.:P __.,a□) provide explicit markers for downstream tasks.

## 5.3. Discrimination Related Vocabulary

Application of the discrimination lexicon (with racism overlap removed) identified 100 distinct surface forms with 466,227 total occurrences. Table 5 lists the 15 most frequent terms, representing 90.8% of discrimination vocabulary.

| Rank | Term (AR) | Gloss (EN) | Frequency | % |
|---|---|---|---|---|
| 1 | _::::J.□il | discrimination | 330,401 | 70.8 |
| 2 | ةاوL-..Jl | equality | 21,539 | 4.6 |
| 3 | --.,; _::::-; | discrimination against | 18,343 | 3.9 |
| 4 | _::::-; | discrimination | 15,873 | 3.4 |
| 5 | _::::J.□ilو | and discrimination | 6,856 | 1.5 |
| 6 | ةاوL-..Jlو | and equality | 6,166 | 1.3 |
| 7 | ي_::::J.□il | discriminatory (adj.) | 4,375 | 0.9 |
| 8 | □: _::l□l | discriminatory (adj.) | 3,518 | 0.8 |
| 9 | _::::_::ll | for discrimination | 3,298 | 0.7 |
| 10 | ث: _..::ll | segregation | 2,854 | 0.6 |
| 11 | □: ..i□Jl | regionalism | 2,582 | 0.6 |
| 12 | ةاوL-,. | equality | 2,294 | 0.5 |
| 13 | نL,.෴ | deprivation | 1,906 | 0.4 |
| 14 | □□ع | injustice | 1,719 | 0.4 |
| 15 | ي..i□Jl | regional (adj.) | 1,682 | 0.4 |
| *Top 15 subtotal* | | | *423,406* | *90.8* |
| *All 100 terms total* | | | *466,227* | *100.0* |

Table 5: Top 15 discrimination-related surface forms with English glosses. Percentages are within the discrimination vocabulary.

The term _::::J.□il (discrimination) dominates at 330,401 occurrences (70.8%), establishing it as the primary lexical marker. Equality related terms (ةاوL-..Jl, ةاوL-,.; 23,833 combined, 5.1%) indicate discrimination is frequently framed through an equality lens, often in negated constructions. The phrase --.,; _::::-; (18,343 occurrences, 3.9%) explicitly marks discrimination targets. The lexicon distinguishes structural discrimination (.s..Jj..Jl, □::lS:::.btl) from interpersonal bias (_::...::ll, ةL□L>--1l). Regional discrimination is captured via □:. ..i□Jl/ي..i□Jl (4,264 combined, 0.9%).

## 5.4. Discrimination Axes

We identified 20 discrimination axes (identity characteristics serving as grounds for differential treatment) with 104,812 total mentions. Table 6 presents the top 10, representing 86.6% of axis mentions.

The 20 discrimination axes were selected based on two criteria: (1) identity-based characteristics that commonly appear as grounds for unequal treatment in Arabic public discourse, and (2) suf-ficient frequency in our corpus to support reliable matching and analysis. We began from protected-characteristic groupings commonly used in discrimination research, then refined and merged categories based on how identity is expressed in Arabic Facebook posts (including dialectal usage).

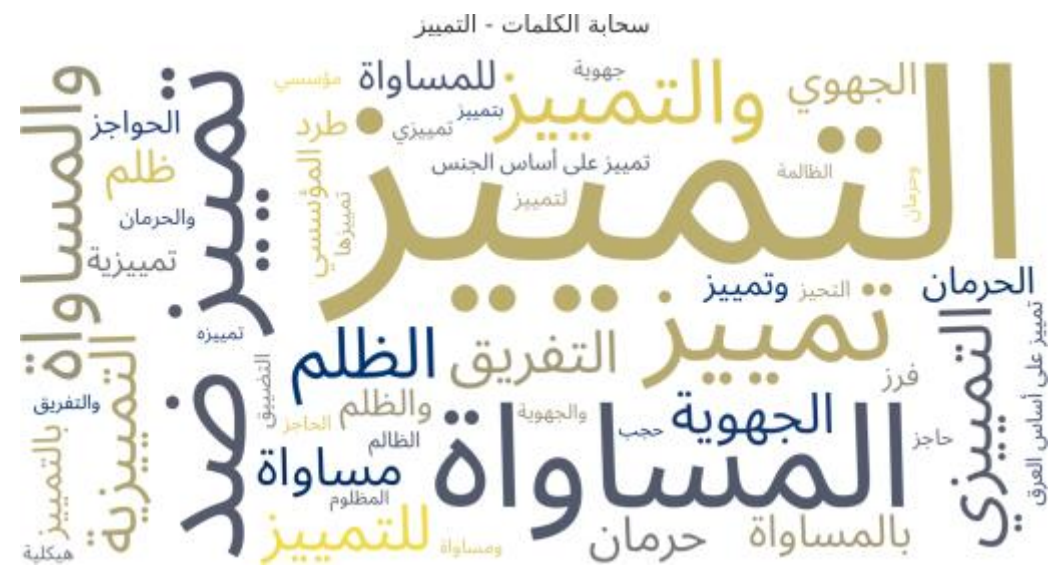

Figure 2: Discrimination related word cloud (top 40 terms). _::::J.□i' is visually prominent; colormap is colorblind safe.

| Rank | Axis (AR) | Gloss (EN) | Frequency | % |
|---|---|---|---|---|
| 1 | □_Il' | language | 17,131 | 16.3 |
| 2 | □Ab..:..J' | region | 11,448 | 10.9 |
| 3 | □::-.i□J' | nationality | 11,277 | 10.8 |
| 4 | □;Lc□I' | disability | 9,328 | 8.9 |
| 5 | الـلّـن' | skin color | 7,882 | 7.5 |
| 6 | I..)".i□J' | gender | 7,671 | 7.3 |
| 7 | __c\I' | origin/ancestry | 7,597 | 7.2 |
| 8 | (;□٪ۆ≤ﻫ٩I' | refugees | 6,354 | 6.1 |
| 9 | __..I' | age | 6,114 | 5.8 |
| 10 | 0-I' | age (variant) | 6,011 | 5.7 |
| *Top 10 subtotal* | | | *90,813* | *86.6* |
| *All 20 axes total* | | | *104,812* | *100.0* |

Table 6: Top 10 discrimination axes with English glosses. Percentages show the proportion within the axes category.

Language (□_Il') is the dominant axis at 17,131 mentions (16.3%), reflecting discrimination based on dialect and multilingualism. Geography adjacent axes (□Ab..:..J', □::-.i□J') represent 22,725 mentions (21.7%), often linked to migration and refugee discourse. Substantial targeting of displacement affected populations (refugees and migrants) totals over 10,000 mentions (9.5% of axes). Race/color based discrimination accumulates to 15,479 mentions (14.8%), providing empirical evidence of racialized discrimination in Arabic Facebook discourse.

## 5.5. Contextual Attribution Patterns

Explicit discrimination attributions following J╗c ر..>,. + ½□-□/سLJأ ("on the basis of + [axis]") were extracted. These deterministic constructions unambiguously link treatment to identity characteristics and serve as high precision cues for discriminatory framing, enabling targeted sampling and annotation bootstrapping. Patterns are provided in supplementary materials with frequency counts.

## 5.6. Resource Utility

The resource supports automatic annotation for downstream classification and axis-aware sampling for targeted linguistic or sociological analysis. Morphology-aware families improve high-recall retrieval at Facebook scale, and we release lexicons, patterns, and scripts with documentation for reproducible use.

# 6. Discussion

Our analysis suggests that discrimination discourse on Arabic Facebook is often expressed through a mix of explicit identity references and broader framing around rights, belonging, and social status. The term _::::J.□i' is consistently prominent, and its surrounding vocabulary frequently includes equality- and fairness-related expressions. This implies that discrimination is not always framed as direct hostility, but also as denial of equal treatment or normalized exclusion. For downstream tasks, this motivates covering both explicit insults and policy-like discourse that legitimizes unequal treatment.

**Racism discourse and sectarian overlap.** The racism lexicon concentrates around a small set of highly productive roots and their dialectal variants, while sectarian or identity cues appear in close proximity in many contexts. This overlap matters for Arabic NLP because the same post can combine racial labeling with other identity-based targeting. Axis-aware sampling is therefore useful, as it allows researchers to isolate specific targets (e.g., nationality or religion) while still capturing mixed framing that reflects real discourse.

**Axes that emerge in practice.** Across the decade-long timeline, language and nationality-related targeting appears frequently, alongside geography- and migration-linked discrimination. These results align with the sociopolitical context of the region and indicate that fairness auditing should not focus on a single axis. Instead, evaluation should cover a range of identity categories that are salient in Arabic discourse, including race/color, religion, and nationality.

**Attribution patterns as high-precision cues.** Explicit attribution structures (e.g., + سLJأ J╗c

[axis]) provide high-precision signals for identifying discrimination framing. While they do not cover all cases, they are valuable anchors for sampling, weak supervision, and qualitative studies, especially when combined with lexicon matches. In practice, these patterns can help build cleaner subsets for annotation and model evaluation, and can support targeted studies of how discrimination is justified or normalized.

**Role of platform signals.** A key added value of ArabDiscrim is that it keeps platform-native engagement signals alongside text. These signals do not automatically imply harm, but they enable questions that text-only corpora cannot answer, such as whether different axes trigger different reaction profiles, or whether certain framing styles are associated with higher shares and comment activity. We view these fields as enabling research on the ecology of discrimination discourse, rather than as direct labels of harmfulness.

## 7. Social Impact and Applications

ArabDiscrim can support multiple stakeholders studying discrimination in Arabic public discourse, while also raising dual-use concerns that require careful handling.

**Research use cases.** First, the lexicons, morphology-aware regex families, and attribution patterns can support weak supervision and bootstrapping pipelines for discrimination detection, especially in dialectal settings where surface forms vary widely. Second, axis-aware sampling enables targeted studies that focus on a specific identity category (e.g., nationality, language, religion), which is important for fairness evaluation and for comparing how different groups are discussed. Third, the decade-long time span enables longitudinal analyses, for example tracking how targeting intensity changes around major sociopolitical events or policy shifts.

**Platform-aware analysis.** By including reactions, shares, and comments, the dataset supports joint analysis of language and audience response. This can help researchers study how discriminatory framing spreads and how communities interact with such content. These fields should be used as contextual signals rather than direct indicators of harmfulness, but they can guide prioritization for qualitative review or for selecting case studies.

**Broader social value.** For civil society and policy-oriented work, ArabDiscrim can help document narratives around migrants, refugees, identity, and belonging in Arabic online spaces. It can also contribute to building evaluation sets that better reflect the linguistic and cultural realities of Arabic discrimination discourse, which remains underrepresented in widely used NLP benchmarks.

**Risks and responsible use.** A dataset and lexicon about discrimination can be misused to amplify harmful language or to target communities. We mitigate this risk by adopting a research-focused access policy, encouraging secure storage, and providing documentation that emphasizes harm reduction. We also encourage users to avoid releasing identifiable information, to report results carefully, and to consider ethical implications when deploying models trained on the resource.

## 8. Conclusion

We introduced ArabDiscrim, a decade-long lexical resource and corpus of 293,056 public Arabic Facebook posts (2014–2024) discussing racism and discrimination. Compared to Twitter-centric resources, ArabDiscrim preserves platform-native engagement signals together with multi-channel textual content, enabling analysis that connects discriminatory framing with how audiences respond and interact.

ArabDiscrim includes (i) 200 curated discrimination-related terms split across racism and discrimination, (ii) morphology-aware regex families to capture inflectional and orthographic variation, (iii) 20 discrimination axes grounded in identity-based targeting, and (iv) explicit attribution patterns that provide high-precision cues for sampling and weak supervision. Together, these components support downstream tasks such as axis-aware corpus sampling, linguistic analysis of discrimination discourse, and platform-aware auditing.

We will release the resource with documentation and supporting scripts under a research-focused access policy aligned with platform terms and ethical considerations. Future work includes expanding dialectal coverage, creating a manually annotated subset for benchmarking, and studying how engagement profiles vary across axes, framing strategies, and time.

## Limitations

The design and scope of ArabDiscrim present several limitations that future research should aim to address:

1. **Lexical Based Retrieval:** The corpus is constructed through lexicon-driven matching. This method, while effective for building a

large scale, topically relevant resource, has inherent limitations in precision and recall.

- **False Negatives (Recall):** The dataset will not capture implicit discrimination, coded language, sarcasm, or discriminatory content expressed through terms not included in the 200 curated lexicons. While morphological expansion captured many inflections, novel slurs or highly localized dialectal expressions may be missed.
- **False Positives (Precision):** The lexicons are designed to retrieve *discourse about* racism and discrimination, not exclusively *performative* hate speech. Consequently, the corpus contains a mix of content, including news reports *on* discrimination, anti racism advocacy, and academic discussions, alongside directly discriminatory posts. The inclusion of terms like ”equality” (ةاوL-,.) highlights this breadth.

2. **Absence of Manual Annotations:** The resource is provided as a large scale corpus with platform signals and is intended for applications like automatic annotation and axis-aware sampling. It is not a gold standard, human annotated dataset for post level classification. As noted in the related work, existing resources (e.g., ADHAR, So Hateful!) fulfill that role, though on a different platform. Users requiring high precision post level labels (e.g., ”hate” vs. ”non hate”) will need to conduct further annotation, as suggested in our future work.

3. **Platform Specific Scope:** By design, this work addresses the gap in Facebook centric resources. This focus, however, means the findings and lexicons are not necessarily generalizable to other platforms like Twitter, TikTok, or Telegram, which have different affordances, user bases, and discourse norms.

4. **Data Source Limitations:** The data was collected from *public* Facebook pages. This excludes discourse happening in private groups or on individual user profiles, where discrimination may be expressed differently or more overtly.

5. **Text Centric Analysis:** The resource primarily captures textual data across four channels (Message, Description, Image Text, Link Text). While this includes OCR’d text from images, it does not fully address multimodal discrimination (e.g., meaning derived from images, videos, or memes).

## Ethics Statement

The construction and release of a dataset focused on racism and discrimination require careful ethical consideration.

1. **Data Sourcing and Privacy:** All data included in ArabDiscrim was collected from public Facebook pages. No data was collected from private user profiles or closed/private groups. The dataset focuses on the *posts* from these pages, not on individual user comments or interactions. While the posts are public, their aggregation into a large scale, queryable dataset presents a re identification risk. The dataset will be released with page level metadata (e.g., Page Name, Page Id) to support platform ecology research and reproducibility, but this requires researchers to handle the data responsibly.

2. **Content and Researcher Harm:** The dataset contains extensive examples of *discourse about* discrimination, racism, and sectarianism. This content is sensitive and often offensive, hateful, and psychologically distressing. Researchers and annotators interacting with this data are at risk of psychological harm from repeated exposure.

3. **Potential for Misuse (Dual Use):** A significant ethical risk is the ”dual use” potential of the resource. Malicious actors could misuse the curated lexicons or the dataset to refine hate speech generation models, identify new targets for harassment, or optimize the spread of discriminatory content.

4. **Mitigation and Justification:** We believe the potential benefits of this resource for combating discrimination outweigh the risks. The dataset is intended to provide a foundation for more ecologically valid research on discrimination, enabling harm mitigation at scale.
   - To mitigate risks, the lexicons, patterns, and dataset will be released exclusively for non commercial research purposes under a restrictive license.
   - Prospective users will be required to submit an application detailing their research aims and ethical protocols to ensure the resource is used for ”bona fide” research aligned with the project’s goals.
   - The inclusion of native Arabic speakers in the validation process helps ensure that linguistic and cultural contexts are appropriately handled, reducing the risk of misinterpreting and mislabeling sensitive discourse.

- We strongly urge researchers using this data to store it securely, restrict access, and adhere to ethical best practices for handling sensitive human data.

## Acknowledgments

This work was made possible by the National Priorities Research Program (NPRP) grant NPRP14C-0916-210015 from the Qatar National Research Fund (QNRF), a member of the Qatar Research, Development and Innovation Council (QRDI).